\documentclass{article}

\usepackage{arxiv}
\usepackage{chngcntr}
\usepackage[utf8]{inputenc} % allow utf-8 input
\usepackage[T1]{fontenc}    % use 8-bit T1 fonts
\usepackage{hyperref}       % hyperlinks
\usepackage{url}            % simple URL typesetting
\usepackage{booktabs}       % professional-quality tables
\usepackage{amsfonts}       % blackboard math symbols
\usepackage{nicefrac}       % compact symbols for 1/2, etc.
\usepackage{microtype}      % microtypography
\usepackage{xcolor}         % color support for \textcolor
\usepackage{lipsum}
\usepackage{graphicx}
\usepackage{amsmath}
\usepackage{threeparttable}
\usepackage{float}
\usepackage{lineno}
\usepackage{xcolor}
\graphicspath{ {./images/} }

\title{Visual Enumeration Remains Challenging \\ for Multimodal Generative AI}

\author{
 Alberto Testolin \\
  Department of General Psychology and \\
  Department of Mathematics \\
  University of Padova, Padova (IT) \\
  \texttt{alberto.testolin@unipd.it} \\
  \And
 Kuinan Hou \\
  Department of General Psychology \\
  University of Padova, Padova (IT) \\
  \texttt{kuinan.hou@phd.unipd.it} \\
  \And
 Marco Zorzi \\
  Department of General Psychology and Padova Neuroscience Center \\
  University of Padova, Padova (IT) and \\
  IRCSS San Camillo Hospital, Venice-Lido (IT) \\
  \texttt{marco.zorzi@unipd.it} \\
}

\begin{document}
\maketitle
\begin{abstract}

Many animal species can approximately judge the number of objects in a visual scene at a single glance, and humans can further determine the exact cardinality of a set by deploying systematic counting procedures. In contrast, it has been observed that even state-of-the-art AI systems have very limited enumeration skills. In this work, we propose two benchmark tasks inspired by cognitive science that allow to precisely evaluate the visual enumeration capabilities of multimodal foundation models, thereby providing an objective measure of their number sense and counting level. We consider popular visual question answering models (BLIP, LLaVA and ViLT) as well as advanced image-to-text (Gemini, GPT and Qwen) and text-to-image (DALL-E, FLUX and Stable Diffusion) AI systems. Our analyses show that even the most advanced models cannot reliably name the number of objects in simple visual stimuli or generate images containing a target number of items, as indexed by their low accuracy in both types of tasks. Especially for numbers outside the subitizing range, their responses are often far from the target numerosity, and, in stark contrast with human behavior, in many cases the distribution of errors depends on the object category. We also observe some striking mistakes  with small numbers. Our findings demonstrate that developing an intuitive visual understanding of number remains challenging for AI models and that merely increasing model size might not be a viable strategy to promote the emergence of systematic counting skills. We release the full code of our benchmark to facilitate the evaluation of enumeration skills in future AI systems.

\end{abstract}


\section{Introduction}

Artificial Intelligence (AI) is progressing rapidly, with deep learning models approaching or even surpassing human performance in a variety of domains, including perceptual judgements \cite{mckinney2020international} and natural language processing \cite{gilardi2023chatgpt}.
However, even the most advanced multimodal AI systems still struggle in judging the numerosity of visual sets, a core cognitive capability that humans share with many animal species \cite{dehaene2011number}. 
Even infants are sensitive to numerosity \cite{izard2009newborn} and toddlers can generate sets that contain a target number of items \cite{sella2016spontaneous}, suggesting that a pre-verbal understanding of numerical quantities develops well before language development and formal education. Small numerosities in the ``subitizing range'' (up to 4) are perceived in an exact manner (i.e., enumeration is error-free), while the numerosity of larger sets is only approximately estimated when counting is precluded \cite{feigenson2004core}. In the latter case, responses follow Weber's law, so that variability increases proportionally to the mean estimate \cite{dehaene2011number}.
Another key signature of our number sense is its abstract nature: similar response patterns are observed for items of all categories, despite variation in object features (such as color or shape). Indeed, numerosity is spontaneously extracted by our visual system \cite{cicchini2016spontaneous} and it is encoded independently from object category, location, or presentation modality \cite{izard2009newborn}.  Moreover, during childhood, humans (but no other species) learn counting algorithms that allow them to establish the exact cardinality of any set of objects by performing a one-to-one mapping between visual or auditory items and the list of number words \cite{gallistel1992preverbal}. Importantly, there is a broad consensus that number sense and mastery of counting principles are foundational for the development of numeracy and the acquisition of higher-level mathematical competence \cite{halberda2008individual,dehaene2011number, carey2019ontogenetic,dolfi2024weaker}.

AI researchers have engineered a variety of specialized computer vision architectures to count objects in visual scenes, often tailored to specific categories such as animals \cite{arteta2016counting}, crowds \cite{khan2023revisiting}  {or common objects encountered in a specific domain of interest \cite{gao2024nwpu}}. The most popular framework consists of running an object detector to first segment the target items in the image and then explicitly counting the resulting bounding boxes or object proposals \cite{trott2017interpretable,zhang2018learning}, in some cases summing fractional counts estimated from different sections of the image \cite{chattopadhyay2017counting}. However, in these approaches numerosity representations do not emerge within the model itself because the encoding of numbers is delegated to an external, hard-wired (and often category-specific) mechanism.
A radically different perspective considers the possibility that numerosity representations might spontaneously emerge in neural systems as a high-order statistical feature of the sensory signal \cite{zorzi2018emergentist}. Indeed, a rudimentary visual number sense has been shown to emerge in small-scale generative models trained with the goal of reconstructing images with a varying number of items \cite{stoianov2012emergence,testolin2020visual} and number-selective neurons have been observed in generic convolutional networks trained for object recognition \cite{nasr2019number,mistry2023learning}. Thus, one might wonder whether a similar capacity could emerge in modern multimodal foundation models \cite{li2024multimodal}, which are large-scale generative architectures trained on huge data sets that exhibit emergent abilities \cite{wei2022emergent} and can readily solve a wide range of downstream tasks \cite{bommasani2021opportunities,bubeck2023sparks}.
Unlike domain-specific architectures engineered for visual counting \cite{huang2024point, shi2024training}, foundation models are domain-general systems that can be used out-of-the-box without the need of fine-tuning on numerical tasks. However, despite their flexibility and their remarkable performance in a variety of domains \cite{romeo2025artificial}, even the most advanced foundation models fall short in tasks that require the manipulation of numerical information \cite{testolin2023can,rane2024can,testolin2024visual}, calling for a systematic investigation of their basic visual enumeration skills.

In line with the proposal of using methods from cognitive science to test AI models \cite{binz2023using}, in this work we address this problem by introducing two benchmark tasks that are commonly used to evaluate enumeration skills in humans: numerosity naming \cite{revkin2008does}, which requires establishing how many items are present in a given stimulus, and numerosity production \cite{whalen1999nonverbal,sella2016spontaneous}, which requires generating a set containing a target number of items. The former task can be used to probe \emph{image-to-text} architectures, while the latter can be used to probe \emph{text-to-image} architectures.
Our benchmark allows to characterize the distribution of model responses using a variety of object categories, providing aggregate scores indicating the overall model performance and additional metrics that measure whether the distribution of responses follows a human-like pattern. Perfect accuracy across the entire numerical range would suggest the emergence of systematic counting skills, while error-free responses with only small numbers would either indicate subitizing capabilities, or that counting is only partially developed as in children who do not fully master the counting principles \cite{le2007one,lee2011number}. Error-prone responses centered on the target number would instead suggest that the AI model relies on numerosity estimation, which may follow Weber's law (as in humans) or not.

We perform our evaluation across a wide range of models of different sizes and complexities,  {considering the most powerful multimodal AI systems and visual question answering models available at the time of the research}.
In the image-to-text domain, we consider AI models that can provide written answers to non-trivial questions about the content of an image or accurate descriptions of complex visual scenes. In particular, we test three popular architectures used in visual question answering: ViLT (vision-and-language transformer) \cite{kim2021vilt}, BLIP (bootstrapping language-image pre-training model) \cite{li2023blip} and LLaVA (large language and vision assistant model) \cite{liu2024visual}. We further test  {Qwen2.5-VL \cite{bai2025qwen2}, which is one of the latest open-source multimodal models available, as well as two proprietary models that are considered among the most advanced multimodal systems currently available}: GPT-4V \cite{yang2023dawn} and Gemini Pro Vision \cite{team2023gemini}.
In the text-to-image domain, we instead consider foundation models that can produce high-quality visual content following detailed user prompts provided in natural language. We test two popular open-source generative architectures for images, Stable Diffusion \cite{rombach2022high,podell2023sdxl} and FLUX \cite{blackforestlabs_flux}, as well as DALL-E \cite{ramesh2022hierarchical,betker2023improving}, which is regarded among the most powerful proprietary systems.  {For Qwen and DALL-E,} we also compare different versions of the same architecture to investigate whether increasing model size supports more refined visual enumeration skills.

 {The research contributions of our work are multifaceted. From a methodological perspective, we introduce a unified experimental procedure to evaluate the numerical skills of both image-to-text and text-to-image AI models, which allows us to quantitatively characterize their numerical competence across different types of object categories. We make our benchmarking pipeline publicly available to allow systematic evaluation of future AI models [\hyperlink{https://github.com/CCNL-UniPD/Numbersense-AI}{https://github.com/CCNL-UniPD/Numbersense-AI}].
From a theoretical perspective, we demonstrate that although larger models generally possess a better number sense, merely scaling-up the model size might not be the best way to spur the emergence of systematic counting skills. In this regard, we show that the weak performance in visual enumeration might be partially due to properties of the training corpora commonly used to build foundation models: for example, in popular training datasets the frequency of numerosities rapidly falls off according to a power law, implying that larger numbers are underrepresented, and textual captions often contain numerical information that is not related to the numerosity of the visual scene, thus injecting noise into the alignment of different input modalities.}

 {The article is structured as follows: in the Materials and Methods section, we describe the problem setting, introducing the benchmark tasks and the evaluation metrics used to quantify numerical competence. We also provide details about the models considered and the relevant baselines. In the Results section, we present the quantitative results comparing different models, along with a detailed analysis of response errors and the statistical properties of two representative corpora commonly used to train large-scale multimodal AI systems. In the Discussion, we review and interpret our findings, while in the Conclusions we highlight their implications for AI research and discuss possible future directions.}

\section{Material and Methods}

\subsection{Benchmark tasks}

\subsubsection{Image-to-text: numerosity naming}
\begin{figure}[t]
\centering
\includegraphics[width=.65\textwidth]{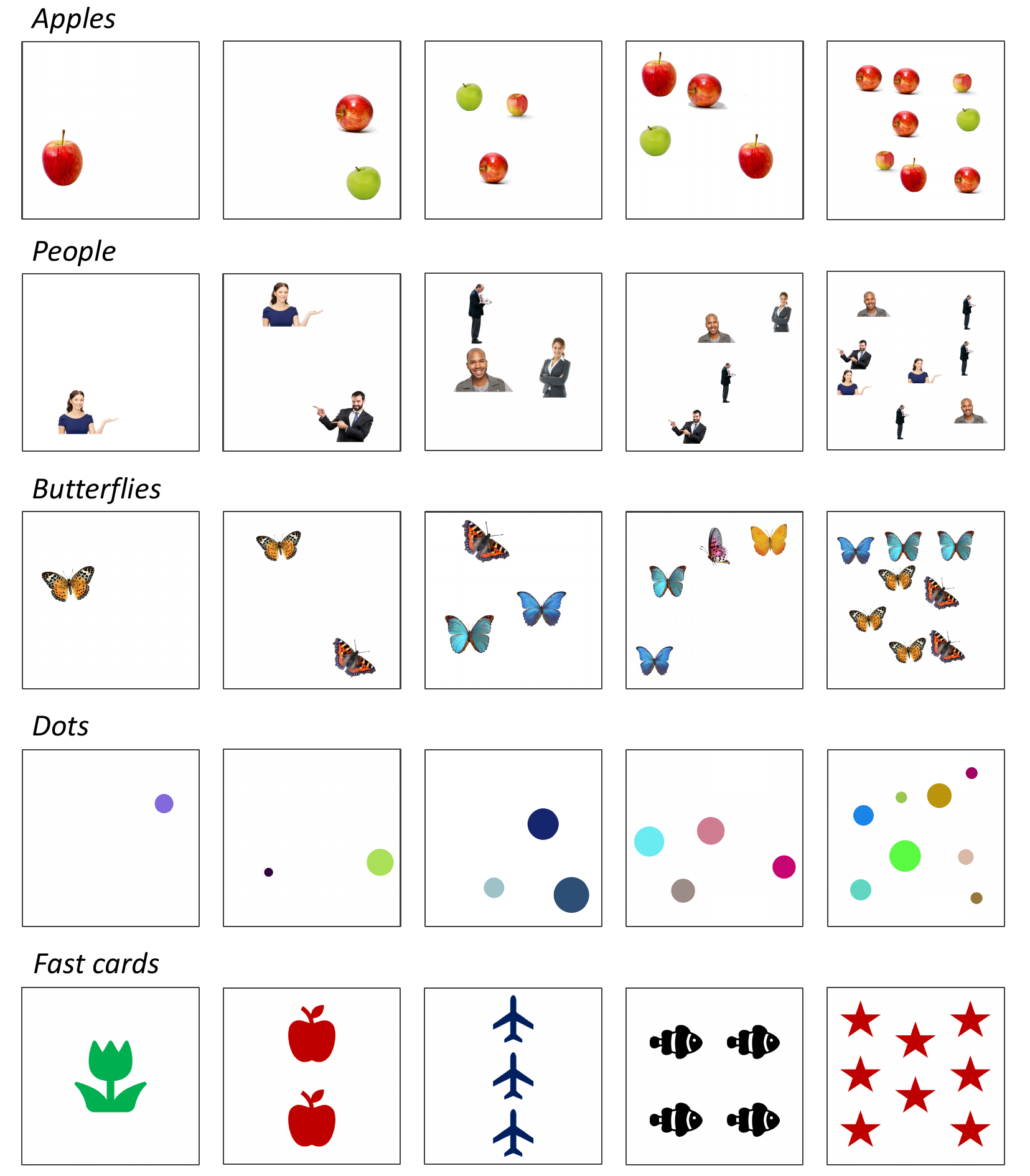}
\caption{Samples from the numerosiy naming task. Each row contains samples from a different object category, while columns correspond to different numerosities: 1, 2, 3, 4 and 8.}
\label{fig:1}
\end{figure}

In the numerosity naming task, the models are asked to establish how many objects are present in a set of simple images containing up to 10 objects. We created a new dataset of synthetic images, each including only items of the same category sampled from 5 possible object types: apples, people, butterflies, colored dots, and ``fast cards'' depicting regularly placed clip-arts similar to those used to test number sense in young children \cite{le2007one}. Examples of stimuli are shown in Figure \ref{fig:1}.
For each object class and target number \textit{n}, we created 50 high-resolution images ($1024\times1024$ pixels) where the \textit{n} objects have variable size and are randomly placed on a uniform white background, with no overlap. Fast card stimuli are created using clip-arts of common objects (apples, bells, butterflies, candies, cars, fish, flowers, planes, stars) drawn in different colors (black, blue, green, orange, red).

For each model tested, we consider three different prompting methods and select the one leading to the best performance, measured as mean absolute distance from the target number. The first prompt requires to explicitly estimate the number of objects belonging to a specific category (i.e., \textit{How many apples / butterflies / people / dots / shapes are there in the picture?}). The other two use the more general words ``objects'' or ``things'' to identify the items to count.
To make sure that the models are prompted correctly, we also perform a control simulation related to a non-numerical task using the entire set of ``apples'' stimuli, probing the models with the following prompt: \textit{What does the image represent?} and considering as correct the following answers: \textit{apple(s)} and \textit{fruit}. All models provided the correct answer for the entire set of stimuli in the control task, thus demonstrating a proper understanding of the image content and the prompt structure.

Model responses are automatically parsed: if present, number words are converted to numerical values using the \href{https://pypi.org/project/word2number/}{word2number} Python library, and responses are discarded if they contain multiple numbers or vague quantification terms (e.g., ``a few'', ``a bunch of''). We verified that at least 20 eligible trials were recorded for each number / object category combination.

\subsubsection{Text-to-image: numerosity production}
In the numerosity production task, each model is asked to generate 100 high-resolution images containing a target number of objects, in analogy with numerosity production tasks used in animal and human studies \cite{whalen1999nonverbal,sella2016spontaneous}. Target objects belong to the same classes used for the naming task, except for the ``fast cards'' category, which could be underrepresented in the corpora used to train foundation models.

All models were initially prompted with the following text: \textit{An image with n apples / butterflies / people / dots} (where \textit{n} varied between 1 and 10). When \textit{n} = 1 the prompt was adjusted to the singular form.
However, for the dots category this prompting method resulted in poor generations: we obtained better results when the models were prompted with a more specific description of the image: \textit{n filled dots in white background}. For the people category, instead, we obtained better results with the prompt \textit{An image with n persons}.

\begin{figure}[t]
\centering
\includegraphics[width=.99\textwidth]{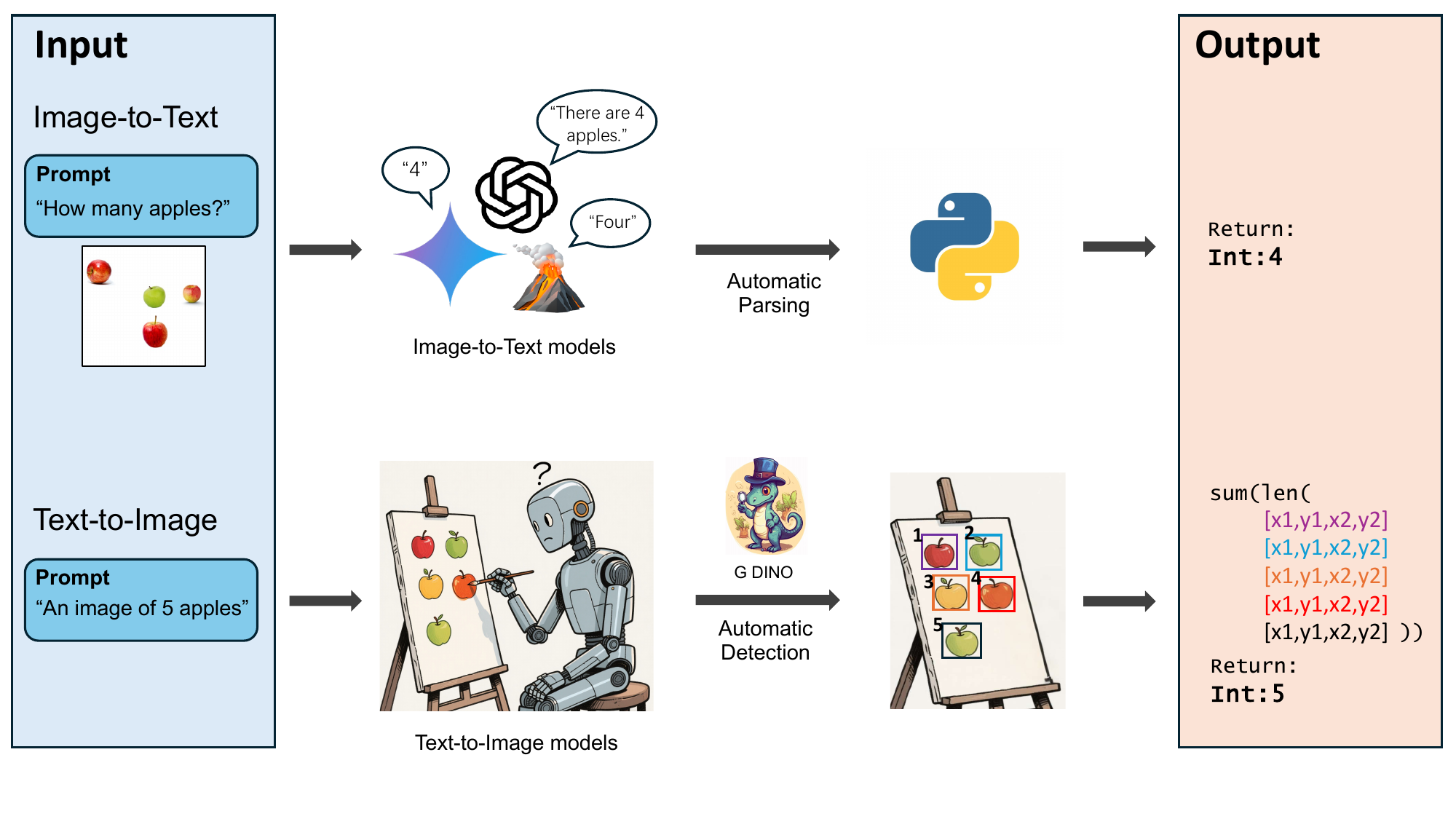}
\caption{ {Graphical representation of our evaluation pipeline. Numerosity naming (image-to-text) is represented in the upper stream, while numerosity generation (text-to-image) is represented in the lower stream.}}
\label{fig:pipeline}
\end{figure}

The generated images are automatically parsed using a computer vision pipeline \cite{hou2025estimating} optimized on a set of human-annotated images (see Supplementary Information for details), which returns the number of objects generated  {(see Figure \ref{fig:pipeline})}. If the automatic pipeline does not find any countable object, the image is discarded and the model is prompted again.

\subsection{Multimodal AI architectures}
For the numerosity naming task, we consider several representative image-to-text foundation models. We first test three popular multimodal architectures used in visual question answering: ViLT \cite{kim2021vilt}, BLIP-2 \cite{li2023blip}, and LLaVA \cite{liu2024visual}. We further consider three large-scale multimodal language models representing the state-of-the-art in AI research:  {the open-source multimodal Qwen2.5-VL model recently developed by Alibaba \cite{bai2025qwen2}}, the multimodal GPT-4V model developed by OpenAI \cite{yang2023dawn} and the multimodal Gemini Pro model developed by Google \cite{team2023gemini}.
All these systems have remarkable visual reasoning abilities and can answer non-trivial questions related to image content (e.g., \textit{What does the image represent? What are the feelings of the people in the scene and why?}).
For ViLT, we test the vilt-b32-mlm version available through \href{https://huggingface.co/dandelin/vilt-b32-mlm}{Hugging Face}. This architecture incorporates text embeddings into a Vision Transformer, allowing it to have a minimal design for vision-and-language pre-training and thus speeding-up model training and inference phases. It has a total of 87.4 million parameters.
For BLIP-2, we test the blip2-flan-t5-xl version developed by Salesforce, also available through \href{https://huggingface.co/Salesforce/blip2-flan-t5-xl}{Hugging Face}. This architecture is an improved version of BLIP \cite{li2022blip} that approaches state-of-the-art performance on several challenging benchmarks \cite{liang2024hemm}; we explored all versions of the backbone models except the t5-xxl model (due to GPU memory constraints) and found that the version that used Flan-T5 as a language model yielded the best accuracy. The chosen model version has a total of 4.1 billion parameters.
For LLaVA, we test the 1.6 version available through \href{https://huggingface.co/liuhaotian/llava-v1.6-34b}{Hugging Face}. LLaVa is an open-source end-to-end trained large multimodal model based on the transformer architecture, which combines a vision encoder and the \href{https://huggingface.co/NousResearch/Nous-Hermes-2-Yi-34B}{Nous Hermes 2} large language model for general-purpose visual and language understanding, achieving impressive chat capabilities \cite{liu2024visual}. The chosen model version has a total of 34 billion parameters.
 {Qwen2.5-VL is the latest version of the Qwen model family \cite{bai2025qwen2}. It has been optimized for visual recognition, image reasoning, captioning, and answering general questions about an image, outperforming many open-source and proprietary models on common industry benchmarks. We focus on the most powerful model version, which has a total of 72 billion parameters, but we also consider the smaller architecture with 7 billion parameters to measure performance gains with respect to model size.}
GPT-4V and Gemini Pro are regarded among the most powerful generalist AI systems to date, thanks to their unprecedented ability to understand and process an arbitrary mix of input images and texts. Technical details regarding the underlying architecture and inner working of these models (including engineered modules that might be used to solve specific tasks) have not been revealed; it has been \href{https://the-decoder.com/gpt-4-architecture-datasets-costs-and-more-leaked/}{speculated} that both these models might have more than one trillion parameters.

For the numerosity production task, we consider three different image generation architectures that have proven capable of generating high-quality images following a textual description, also taking into account stylistic instructions, fine-grained details, and relational features (e.g., \textit{A photo of an astronaut riding a horse in photorealistic style}).
One is represented by the open-source Stable Diffusion (SD) model family, developed by Stability AI and freely available through \href{https://huggingface.co/stabilityai/stable-diffusion-3.5-large}{Hugging Face}. It is a latent diffusion model that combines an autoencoder with a diffusion model trained on the latent space of the autoencoder. We test both version 2.1 \cite{rombach2022high}, which has approximately 500 million parameters, and the newest version 3.5 Large, which has approximately 8 billion parameters.
We then consider the open-source FLUX model \cite{blackforestlabs_flux}, which is also a diffusion-based architecture with a hybrid design that combines multimodal and parallel diffusion transformer blocks, allowing for a more effective processing of visual and textual data. FLUX has approximately 12 billion parameters, providing enhanced capacity for generating high-resolution, hyper-realistic images and accurately rendering complex visual scenes and text.
Finally, we test two proprietary systems from the DALL-E model family using the API interface provided by OpenAI. \href{https://openai.com/research/dall-e}{DALL-E 2} \cite{ramesh2022hierarchical} is an improved version of the original text-to-image DALL-E model, featuring a total of 3.5 billion parameters. \href{https://openai.com/dall-e-3}{DALL-E 3} \cite{betker2023improving} is the latest and most powerful version, which was trained using highly descriptive synthetic captions for the training images. Its number of parameters is currently unknown.

 {
\subsection{Counting-specific baselines}
As baselines for the numerosity naming task we also evaluate two state-of-the-art architectures specifically tailored for counting tasks: the Point, Segment, and Count (PseCo) model \cite{huang2024point}, which is a detection-based counting model that utilizes point-level supervision and segmentation cues to improve object localization and enumeration, and the Training-Free Object Counting model \cite{shi2024training}, which can perform category-agnostic object counting without additional training, leveraging pre-trained feature extractors.}

\subsection{Evaluation metrics}

\subsubsection{Overall performance score}
For each benchmark task, in addition to accuracy, we also compute the Normalized Absolute Error (NAE) score, which is a commonly used metric for evaluating counting abilities that addresses limitations of the most commonly used Mean Absolute Error (MAE). Indeed, MAE treats all errors equally, while NAE normalizes the absolute error by dividing it by the target value, making it sensitive to proportional rather than absolute differences. This normalization ensures fairness across scales, as larger targets inherently permit greater absolute errors without compromising accuracy. NAE also aligns with perceptual principles like Weber's law, reflecting the proportional nature of human numerosity estimation. NAE is defined as:

\[
\text{NAE} = \frac{1}{n} \sum_{i=1}^{n} \frac{|G_i - T_i|}{T_i}
\]

where \(n\) is the total number of test samples, \(G_i\) is the generated numerosity for the \(i\)-th test sample and \(T_i\) is the target numerosity for the \(i\)-th test sample.

As a baseline, we report the NAE of a random model probed on the same number of test trials. This model generates responses randomly sampled from a uniform distribution in the range 1-20 (see related confusion matrix in Supplementary Figure \ref{fig:base_human_CM}), obtaining a NAE of 2.38.

\subsubsection{Counting level}
We also assess the counting level of each model by applying standard criteria used in the literature on the development of counting skills \cite{le2007one}. To be considered an ``$n$-knower'' (i.e.,  ``1-knower'', ``2-knower'', ``3-knower'', ``4-knower'') the model has to:
1) correctly return $n$ at least 67\% of the time when tested for $n$; and
2) return $n$ less than 50\% of the time when the target number is different from $n$. The counting level is assessed on the average responses collected across all object categories.

\subsubsection{Estimation performance}
When explicit counting is precluded, in humans and other animal species numerosity estimation tasks yield a distribution of errors that varies systematically according to Weber's law \cite{shepard1975internal}. In particular, while responses are error-free in the 1-4 subitizing range, for larger numerosities the standard error of the estimates increases proportionally to the mean, indicating scalar variability \cite{dehaene2003neural,gallistel2000non}.
To investigate whether the AI responses follow a human-like estimation pattern, we measure the Pearson correlation between the confusion matrices produced by the models and that obtained from an ideal human observer (see related confusion matrix in Supplementary Figure \ref{fig:base_human_CM}) that estimates numerosity in accordance with Weber's law, assuming a standard Weber fraction $w$ of 0.15 \cite{piazza2010developmental} and error-free responses in the subitizing range.

 {
\subsubsection{Over- and under-estimation trends}
To investigate the presence of systematic biases in visual enumeration, we analyzed the distribution of prediction errors (i.e., response - target), separately for each model and category. We applied the Wilcoxon signed rank test to assess whether prediction errors significantly deviated from zero (with alpha level set to 0.05). This non-parametric test, chosen for its robustness to non-normal distributions, involves ranking the absolute values of the errors, then comparing the sum of ranks for positive errors against the sum of ranks for negative errors. We then computed the effect size ($r$) using Cohen's formula:}

\begingroup
\begin{equation}
r = \frac{|z|}{\sqrt{n}}
\end{equation}
\endgroup

{where $z$ is the z-score derived from the Wilcoxon test statistic and $n$ is the sample size. Following standard practice, we interpret $r \geq 0.3$ as evidence of a meaningful bias (medium-to-large effect size): a model is classified as overestimating if the sum of ranks for positive errors significantly exceeds that of negative errors and $r \geq 0.3$, and as underestimating if the opposite occurs and $r \geq 0.3$.}

\section{Results}

Table \ref{table_model_performance} summarizes all results from our benchmark, ranking the models in terms of NAE. Despite the simplicity of the enumeration tasks, none of the models achieves perfect accuracy. Counting is particularly poor, with no model exceeding the level of 4 items. In humans, accurate performance up to 4 is also supported by subitizing \cite{feigenson2004core}, which in turn suggests that current multimodal foundation models cannot count at all. Some of the models mimic the pattern of human visual estimation, as indicated by the strong correlations with the confusion matrices produced by the ideal human observer.

\begin{table}[ht]
\centering
\caption{Leader-board according to Normalized Absolute Error (NAE). The Corr w/ Human column reports the correlation with the confusion matrix produced by an ideal human observer. The last column reports the estimated number of model parameters (in Billions).}
\newrobustcmd{\B}{\bfseries}
\begin{tabular}{@{}lcccccc@{}}
\toprule
Model & Accuracy $\uparrow$ & NAE $\downarrow$  & Counting level  & Corr w/ Human   & Size (Billion) \\
\midrule
\multicolumn{6}{l}{\textbf{\textit{Image-to-text models}}} \\
 {Qwen-72B} & \B0.89  & \B0.01 & 4-knower  & 0.89 & 72 & \\
 {Qwen-7B} & 0.82  & 0.02 & 4-knower  & 0.90 & 7 & \\
Gemini Pro  & 0.60  & 0.10   & 4-knower  & 0.93 & 600?  & \\
GPT-4V  & 0.74  & 0.13   & 4-knower  & 0.92 & >1000?  & \\
LLaVa   & 0.37  & 0.13   & 2-knower  & 0.85 & 34 & \\
VILT    & 0.28  & 0.27   & 1-knower  & 0.67 & 0.1 & \\
BLIP2   & 0.29  & 0.33   & 1-knower  & 0.52 & 4  & \\
\multicolumn{6}{l}{\textbf{\textit{Text-to-image models}}}\\
FLUX    &\B0.44  &\B0.25 & 2-knower  & 0.89 & 12  & \\
SD3.5   &  0.41  &  0.26 & 3-knower  & 0.91 & 8   & \\ 
DALLE-2 &  0.34  &  0.37 & 2-knower  & 0.83 & 3.5 & \\
SD2.1   &  0.28  &  0.39 & 1-knower  & 0.77 & 0.5 & \\
DALLE-3 &  0.32  &  0.47 & 1-knower  & 0.84 & ?   & \\
\bottomrule
\end{tabular}
\label{table_model_performance}
\end{table}

\subsection{Image-to-text models}

All models achieved the minimum number of eligible trials required, without the need of further prompting (total number of responses discarded for ViLT: 0; BLIP-2: 6; LLaVA: 0; Qwen: 0; GPT-4V: 0; Gemini: 20). For all models, the best performance was achieved with the generic ``things'' prompt, while for BLIP-2 and LLaVa the best performance was achieved with the category-specific prompts.

\begin{figure}[t!]
\centering
\includegraphics[width=.9\textwidth]{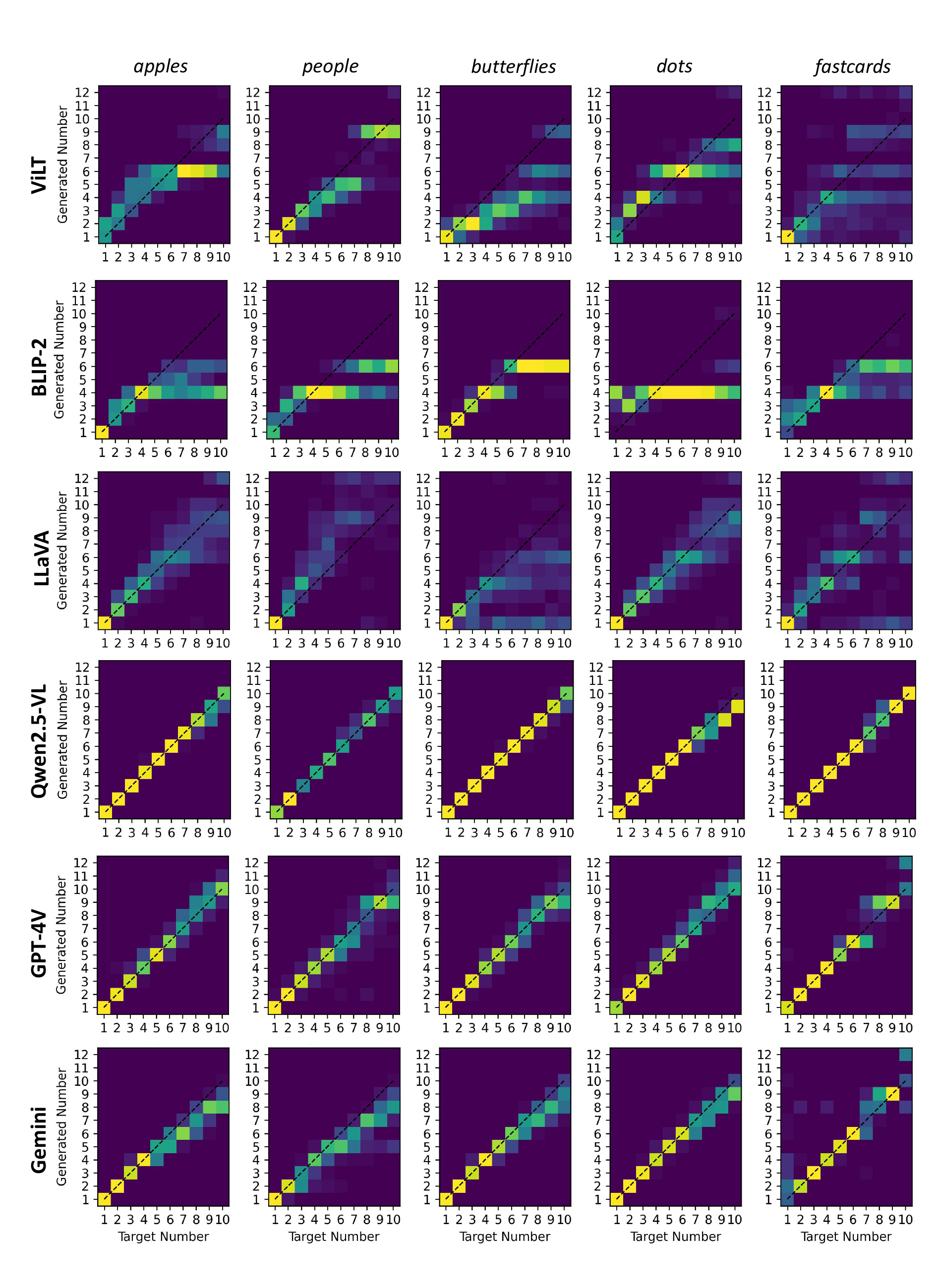}
\caption{Confusion matrices for the numerosity naming task. Each panel shows the distribution of models' responses across different object categories: apples, people, butterflies, dots and fast cards. The x-axis represents the target number, while the y-axis represents the corresponding model responses. Response frequency is encoded using a perceptually uniform colormap (blue = 0\%, yellow = 100\%).  {Qwen2.5-VL stands for Qwen2.5-VL 72B.}}
\label{fig:2}
\end{figure}

For ViLT and BLIP-2 the response accuracy was lower than 30\%. The corresponding confusion matrices (CMs) reported in Figure \ref{fig:2} clearly show the presence of anchoring effects, leading these models to choose stereotyped responses (e.g., 4 or 6).
The pattern of responses for these models also drastically varied between categories (minimum correlation between CMs for ViLT: 0.04, BLIP-2: 0.27), suggesting that they fail to abstract numerical information. Moreover, in sharp contrast with human adults, ViLT and BLIP-2 often returned wrong answers even for images with only one or two objects. According to standard criteria used in human developmental studies, these models can be considered ``1-knowers'', that is, they can only reliably enumerate single objects, as typical of children younger than three years of age \cite{le2007one}. 
The responses of LLaVA were slightly more accurate (37\%) and consistent across object categories (minimum correlation between CMs: 0.52). This model achieves a ``2-knower'' counting level and also shows a more robust correlation with the confusion matrix produced by the ideal human observer.

Among the large multimodal models,  {Qwen achieved the highest accuracy (89\%) with a NAE of 0.01, suggesting that it has the strongest enumeration capabilities. This model performs well across different object categories (minimum correlation between CMs: 0.85) and shows reliable enumeration up to four items, which makes it a ``4-knower''. Interestingly, the smaller version of Qwen with only 7B parameters still performs reasonably well, reaching an accuracy of 82\% and demonstrating a higher correlation with the pattern of responses provided by the ideal human observer.}
The responses of proprietary foundation models, despite their larger size, are generally less accurate than Qwen (GPT-4V: 74\%; Gemini: 60\%), suggesting that these systems possess very rudimentary enumeration skills. Confusion matrices are consistent between categories (the minimum correlation is 0.87 for GPT-4V and 0.70 for Gemini). In some cases Gemini produces unexpected responses with images containing only one item, for example answering that \textit{There are two things in the image: an apple and a white background} (these responses were discarded).

Considering the distribution of response errors, GPT-4V and Gemini, like  {Qwen}, can be characterized as ``4-knowers'', that is, they exhibit reliable enumeration only up to four items. As noted before, this level of performance is consistent with subitizing (i.e., parallel individuation of visual items), but it also highlights the failure in mastering counting skills.

{As shown in Table \ref{tab:over-under_img2txt}, image-to-text models have a general tendency to underestimate the numerosity across all object categories. Qwen-72b and GPT-4V are less biased compared to the other models, showing underestimation (Qwen) or overestimation (GPT) trends only for the Dots category.}

\begin{table}[h]
\centering
\begingroup
\caption{
  Analysis of image-to-text models' estimation biases across object categories. 
  ``-'' indicates no systematic bias, 
  ''Over.'' means the model's responses are systematically higher than the ground truth, and ''Under.'' means they are systematically lower. 
  Numbers in brackets represent the effect sizes (Cohen's $r$). 
}
\begin{tabular}{@{}lccccc@{}}
\toprule
Model & Apples & People & Butterflies & Dots & Fastcards \\
\midrule
ViLT & - & Under. (0.44) & Under. (0.68) & - & - \\
BLIP2 & Under. (0.64) & Under. (0.61) & Under. (0.62) & Under. (0.46) & Under. (0.54) \\
llava34b & - & Over. (0.67) & Under. (0.68) & - & - \\
Qwen-72b & - & - & - & Under. (0.51) & - \\
Qwen-7b & Under. (0.46) & - & Under. (0.33) & Under. (0.40) & - \\
GPT-4V & - & - & - & Over. (0.48) & - \\
Gemini Pro & Under. (0.61) & Under. (0.67) & Under. (0.50) & Under. (0.42) & Over. (0.33) \\
\bottomrule
\label{tab:over-under_img2txt}
\end{tabular}
\endgroup
\end{table}

 {Notably, the counting-specific architectures considered as baseline were not able to successfully count the target objects across all categories, suggesting that models trained with real images might struggle with generalizing their counting skills on synthetic visual stimuli (see Figure \ref{fig:7} in Supplementary Information).}

\subsection{Text-to-image models}

\begin{figure}[t!]
\centering
\includegraphics[width=.99\textwidth]{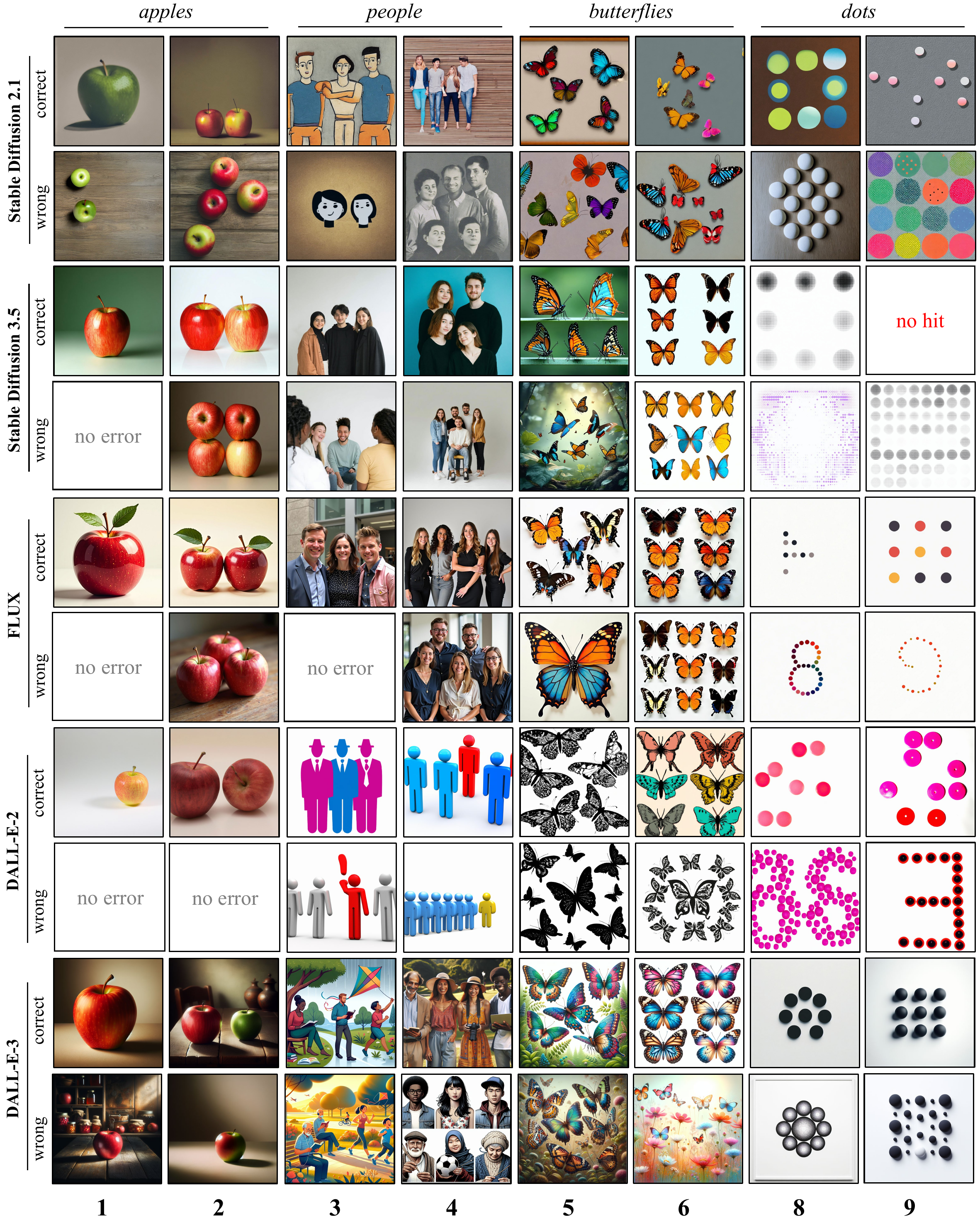}
\caption{Examples of images generated by text-to-image models in the numerosity production task, showcasing both correct and wrong generations (the target number is indicated at the bottom). We report two images for each target category: apples, people, butterflies, and dots. For the dots category, in a few cases FLUX and DALL-E 2 generated images containing a wrong number of dots, which were nevertheless arranged according to the target digit shape (e.g., 8 in the figure). Surprisingly, SD3.5 was unable to generate a single correct response when the target numerosity was larger than 8.}
\label{fig:3}
\end{figure}

Examples of generated images are shown in Fig. \ref{fig:3}. In comparing the outputs of different text-to-image models, we observed noteworthy stylistic trends across both open-source and proprietary systems.
The earlier version of Stable Diffusion (SD2.1) demonstrates a broader artistic scope, spanning paintings, clip-art, and realistic renderings. The latest open-source models (SD3.5 and Flux) appear to have narrowed their stylistic variations, resulting in more coherent and realistic images.
Proprietary models from the DALL-E family instead produce visual scenes with a characteristic synthetic style. DALL-E 2 mostly generates images with white backgrounds and well-defined objects, while DALL-E 3 injects more details, but still retaining a cartoon-like style (in a few cases DALL-E 2 generated images containing reflections: we discarded these trials to avoid ambiguous identification of objects by the automatic evaluation pipeline).

\begin{figure}[t]
\centering
\includegraphics[width=.7\textwidth]{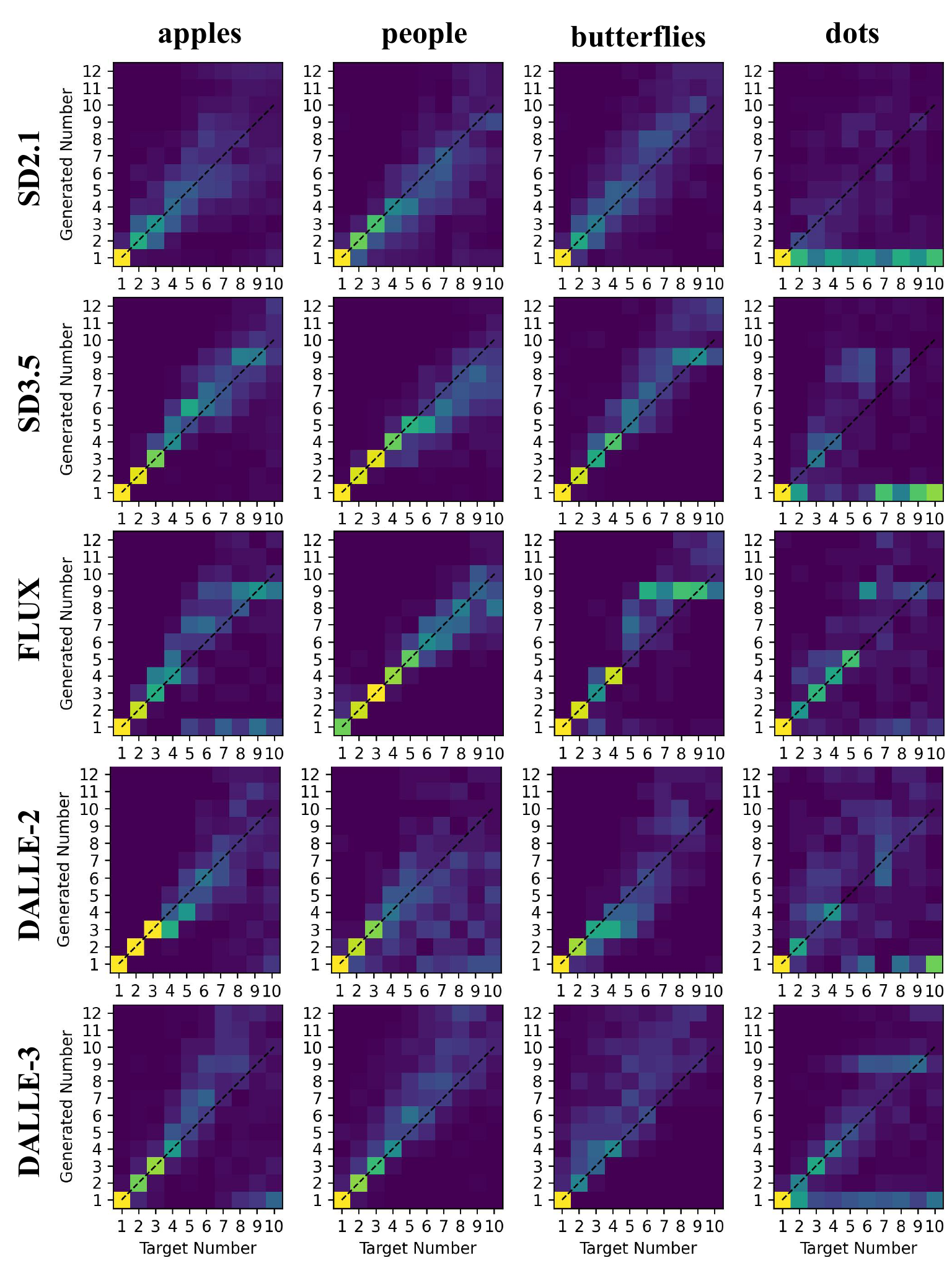}
\caption{Confusion matrices for the numerosity production task. The x-axis represents the target number, while the y-axis represents the corresponding model responses. Response frequency is encoded using a perceptually uniform colormap (blue = 0\%, yellow = 100\%).}
\label{fig:4}
\end{figure}

The mean response accuracy was fairly low for all models (ranging between 0.28 and 0.44), suggesting that the numerosity production task is far more challenging than the numerosity naming task. Confusion matrices are shown in Figure \ref{fig:4}. Similarly to the naming task, the response patterns were not homogeneous across categories (minimum correlation between CMs for SD2.1: 0.43, SD3.5: 0.37, FLUX: 0.69, DALL-E 2: 0.63, DALL-E 3: 0.56).
In a few cases SD3.5, FLUX and DALL-E 2 exhibit error-free responses, but that mostly happens for the generation of a single object (and not across all categories). All other models make errors even in this condition. The NAE ranges from 0.25 (FLUX) to 0.47 (DALL-E 3).
According to criteria used in human developmental studies, the highest counting level is achieved by SD3.5, which can be classified as a ``3-knower''.

 {Table \ref{tab:over-under_txt2img} shows the results of the analysis of over- vs. under-estimation trends for all text-to-image models; in this case we observe a tendency to overestimate.}

\begin{table}[h]
\centering
\begingroup
\caption{Analysis of text-to-image models' estimation biases across object categories. 
  ``-'' indicates no systematic bias, 
  ''Over.'' means the model's outputs are systematically higher than ground truth, and ''Under.'' means they were systematically lower. 
  Numbers in brackets represent the effect sizes (Cohen's $r$).}
\begin{tabular}{@{}lccccc@{}}
\toprule
Model & Apples & People & Butterflies & Dots \\
\midrule
SD2.1 & Over. (0.3) & - & Over. (0.45) & - \\
SD3.5 & - & Under. (0.44) & Over. (0.57) & - \\
FLUX & - & - & Over. (0.45) & - \\
DALLE-2 & - & Under. (0.34) & - & - \\
DALLE-3 & - & Over. (0.64) & Over. (0.76) & - \\
\bottomrule
\label{tab:over-under_txt2img}
\end{tabular}
\endgroup
\end{table}

\subsection{Why is Visual Enumeration so Challenging?}
The poor performance of state-of-the-art AI models in a simple task like visual enumeration might seem surprising, given their impressive range of emergent abilities and considering that signatures of number sense have been observed in smaller-scale deep learning models (for discussion, see \cite{zorzi2018emergentist}).

\begin{figure}[t]
    \centering
    \includegraphics[width=0.85\linewidth]{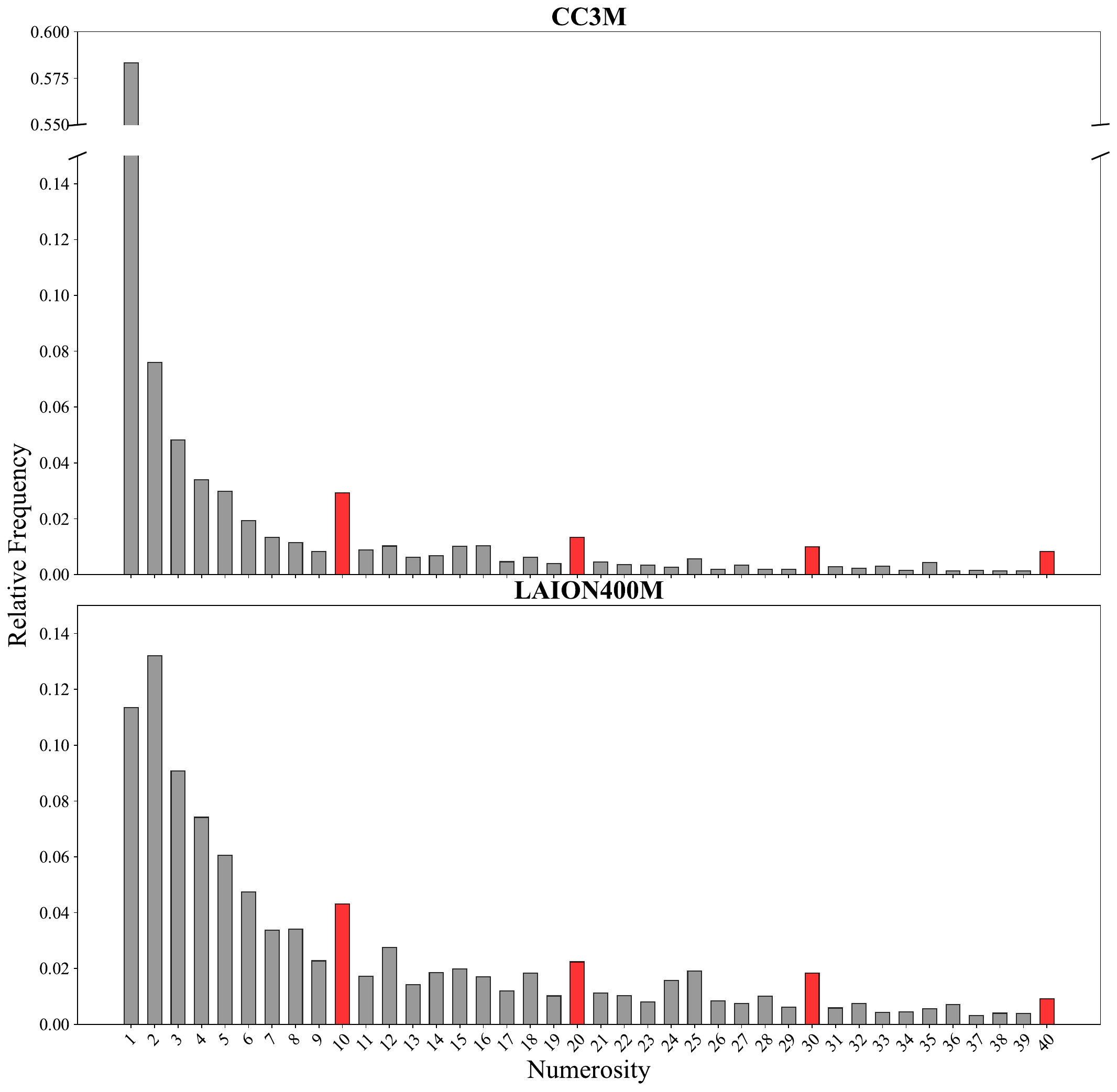}
    \caption{Power-law distribution of textual numerosities related to countable objects in the CC3M dataset (upper panel) and LAION-400M dataset (lower panel), with decade numbers highlighted in red. The y-axis represents the relative frequency of appearance.  {A broken y-axis is used to accommodate the large difference in peak frequencies between the two datasets. The layout is proportioned such that the lower-frequency regions of both datasets share the same vertical height, making them visually comparable despite the scale difference.}}
    \label{fig:num_distri}
\end{figure}

One possible explanation lies in the statistical properties of the material used to train these foundation models. Indeed, it has been recently shown that the performance of multimodal models scales linearly as the concept frequency in pre-training data grows exponentially, which means that ``zero-shot'' performance in tasks involving underrepresented concepts will normally be poor \cite{udandarao2024no}.
To better characterize possible biases in the distribution of numerical information in popular training datasets, we conducted an in-depth analysis on the frequency of appearance of different numerosities in   {two datasets} commonly used to train large-scale multimodal AI systems:   {Conceptual Captions 3 Million \href{https://ai.google.com/research/ConceptualCaptions/download}{(CC3M)} \cite{sharma2018conceptual}, which contains more than 3 million image-caption pairs, and} \href{https://laion.ai/blog/laion-400-open-dataset/}{LAION-400M} \cite{schuhmann2021laion}, a large collection of 400 million English (image, text) pairs. Our investigation focused on examining the distribution of textual numerosities across all image captions present in the datasets, deploying natural language processing tools to identify numbers referring to countable objects in the image (see Supplementary Information for details).  {We analyzed the numerical information in image captions merging together heterogeneous object types, that is, all numerosities mentioned in a caption were summed into a single value, regardless of the object categories. For example, the caption “An image of 3 bananas and 2 apples” is counted as a total numerosity of 5, treating the quantities as a single aggregate and thus capturing the overall number of items described in the caption.} Our analysis revealed that in both datasets numerosities are distributed according to a power law, which implies that larger numerosities are strongly underrepresented compared to small numerosities (see Figure \ref{fig:num_distri}). Interestingly, the frequency of appearance of decades (i.e., 10, 20, 30, 40, etc.) decreases less sharply, indicating a preferential bias for these regular numerosities in the training material.

Such biased distribution of numerosities might contribute to the weak enumeration abilities we observed in all models, especially for images containing a large number of items. Nevertheless, it has been previously shown that signatures of number sense can emerge in smaller-scale models trained with synthetic (black and white) images containing a variable number of geometric shapes, even when numerosities are sampled according to the power-law Zipfian distribution observed in natural images \cite{testolin2020numerosity}. This suggests that the poor number sense of large-scale AI systems might also (at least partially) stem from the more complex visual properties of real images, whose richness in finer-grained details could prevent the emergence of more explicit numerosity representations.

Finally, an additional issue could be the presence of noise in the linguistic captions, since the numerical information provided in the descriptive text sometimes might be completely unrelated to the numerosity depicted in the image. We manually checked a few image-text pairs containing explicit numerical information in their captions to assess whether the reported numerosities accurately matched the corresponding visual content, and indeed we found that sometimes the numerical text is misaligned with the image content (see examples in Supplementary Figure \ref{fig:missalign}). These noisy (image, text) pairs make it challenging to achieve an accurate semantic alignment between visual and textual representations, amplifying problems related to the modality gap observed in multimodal AI systems \cite{liang2022mind}.

\section{Discussion}

The present work demonstrates that modern AI systems cannot yet reliably enumerate the number of objects in a visual scene, both in image-to-text and text-to-image tasks. Such a striking deficit is often observed even for sets containing only a few items, suggesting a counting level that is at best comparable to that of preschool children who do not fully master the counting principles \cite{le2007one,lee2011number}. Exact enumeration up to $n$=4 is consistent with subitizing, which in humans is supported by fast parallel individuation related to object tracking \cite{fu2022electrophysiological} and is independent from counting skills. This observation fits well with the finding that the best performing models generate responses to larger numbers that broadly follow the pattern of human numerical estimation, with scalar variability of the response distribution. Overall, these findings demonstrate that multimodal foundation models do not master counting skills, though the best models exhibit sparks of human-like number sense.

The ability to represent and manipulate visual numerosity should be regarded as a foundational skill for multimodal AI systems because it would ground the subsequent learning of more complex numerical and arithmetic concepts. Numerosity in humans is a primary visual feature (just as orientation or color) \cite{burr2008visual} and it is encoded by neuronal populations in multiple cortical regions in the primate brain \cite{harvey2013topographic,castaldi2019attentional,paul2022numerosity,nieder2016neuronal}. Computational modeling studies have shown that sensitivity to numerosity can indeed emerge in small-scale deep learning models trained to generate synthetic images of object sets \cite{stoianov2012emergence,testolin2020visual,boccato2021learning}: diffusion models, such as Stable Diffusion, FLUX and DALL-E, are trained with a similar objective on huge and heterogeneous image datasets that most likely contain substantial variability in numerosity. However, our analyses have shown that the empirical distribution of numerosities in the (image, text) pairs commonly used to train these systems follows a power law, therefore it might be possible that oversampling of small numerosities in the training corpora of foundation models has detrimental effects on their emergent representational space \cite{udandarao2024no}.
Another issue might lie in the mapping between perceptual numerosity representations, encoded in image embeddings, and number symbols (number words or Arabic digits) encoded in text embeddings. In children, establishing such a bidirectional mapping is a sophisticated developmental process, which takes many years and requires explicit instruction \cite{mundy2009children}. The noisy nature of the (image, text) pairs used to train multimodal models might prevent the creation of a systematic mapping between different input modalities, with detrimental effects on the emergence of abstract numerosity representations.

It is also interesting to observe that even the most advanced proprietary models do not exhibit perfect accuracy on such simple tasks, suggesting that numerosity estimation has not been explicitly built-in in these systems. Nevertheless, one cannot exclude the possibility that some counting mechanisms were at least partially engineered as extra processing layers during prompt elaboration. Furthermore, it is well-known that the most recent AI systems can exploit the self-generation of code snippets to fulfill a user request, as in the case of mathematical problem solving \cite{drori2022neural}, therefore models like GPT or Gemini could in principle also exploit external tools (e.g., based on object detection and a symbolic counting algorithm) to carry out these visual enumeration tasks.
Quite surprisingly, however, in fact we noticed that compared to our own previous investigations \cite{testolin2024visual} the performance of some proprietary models (DALL-E) has slightly degraded, suggesting that rather than pushing for improving number sense, the newest API provided by OpenAI might have actually reduced the compute budget available to the user.
These issues highlight the dangers of using proprietary models in academic research \cite{palmer2024using}, and calls for a broader adoption of open-source alternatives. In the case of our visual enumeration benchmark, it is encouraging to see that the best performance in both numerosity naming and numerosity production tasks was achieved by open-source models (Qwen and FLUX, respectively), demonstrating that future scientific investigations could be carried out by relying on transparent and well-documented neural architectures.

%In this respect, it would be valuable to analyze and compare the embeddings emerging in different models to gain a deeper understanding of the computational principles that could enable the development of stronger enumeration skills.  Unfortunately, the closed-source nature of proprietary systems does not allow us to draw strong conclusions about the nature of their putative visual number sense. Indeed, this would require ``opening the box'' and inspecting the model's inner functioning in a way that is not currently possible with proprietary models. This issue highlights the dangers of using proprietary models in academic research \cite{palmer2024using}, and calls for further research efforts to develop open source foundation models with a focus on basic perceptual abilities, such as those underlying our visual number sense.

\section{Conclusions}

This work shows that systematic visual counting skills do not spontaneously emerge even in the most advanced foundation models, suggesting that significant progress in the design and training of multimodal architectures is still required to create systems that can reliably process visual quantities \cite{testolin2020challenge}.
Efforts to improve the enumeration capabilities of multimodal models have been recently flourishing, highlighting the relevance of this type of benchmark for a comprehensive assessment of their emergent abilities \cite{testolin2024visual,kajic2024evaluating}. For example, recent work suggests that fine-tuning the basic CLIP image-to-text model with a counting-contrastive loss can improve its ability to count the number of objects in images up to ten \cite{paiss2023teaching}.
 {Another  recently proposed approach is to enhance numerical reasoning by improving the numerical captions of the images used in training corpora \cite{jeong2025nucap}.}
For text-to-image models, others have proposed to implement counting guidance by using gradients of a counting network during the generative diffusion process, which however seems effective only for objects with a relatively simple shape \cite{kang2023counting}.
%Another approach tried to first identify features within the diffusion model that can carry information about each single object identity, which are then used during the denoising process to count object instances and eventually steer the process to correct over- or under-generation \cite{binyamin2024make}. Yet another recent work leveraged a dual loop meta-learning framework to create domain-invariant prompts for guiding object quantification, using a pre-trained YOLO object detector to estimate the number of objects generated at each iteration \cite{sun2024quota}.
We believe that our benchmark constitutes an important step forward to investigate these issues, possibly extending the upper limit of tested numerosities well beyond the 1-10 range as the performance of future AI models will improve.

It should also be noted that all models tested in the present work generate their output in a single step, thereby mimicking the kind of processing supported by an approximate estimation system operating in parallel \cite{nieder2004analog}. However, in order to determine the exact number of items in a visual scene humans have learned to deploy iterative counting algorithms, which allow establishing a one-to-one mapping between visual items and number symbols \cite{carey2019ontogenetic}. This nevertheless requires generating outputs in a sequential manner, in analogy to ``chain-of-thought'' methods that have indeed proven useful to improve AI reasoning \cite{wei2022chain}. Whether a similar approach could be used to tackle the enumeration tasks presented in this work is still an open question {, which might be addressed by considering modern methods that can incrementally guide image synthesis and editing using multiple modalities \cite{zhan2023multimodal}}.

In conclusion, we believe that a better visual grounding of numeracy development in AI models could be the key to enabling these systems to acquire and more reliably master mathematical knowledge \cite{mirzadeh2024gsm}  {or even geometrical principles \cite{rudman2025forgotten}} without resorting to highly specialized hybrid architectures \cite{romera2024mathematical}, which could open new possibilities for the use of AI in symbolic reasoning and knowledge discovery \cite{wang2023scientific}.

\section*{Data availability}
All simulations and analyses were implemented using Python v3 and \href{https://colab.research.google.com/?utm_source=scs-index}{Google Colab}. To test larger-scale models (e.g., BLIP-2 or Qwen) we used a virtual machine with an NVIDIA L4 GPU and 64 GB of RAM memory, which was allocated using the Google cloud computing platform.
The data used in the current study and the complete code of our benchmark are freely available on GitHub at: \href{https://github.com/CCNL-UniPD/Numbersense-AI}{https://github.com/CCNL-UniPD/Numbersense-AI}.

\section*{Acknowledgements}
We are grateful to OpenAI for granting research access to the GPT-4V and DALL-E APIs. This work was partially supported by the European Union Next-Generation EU grant (grant n.PE13 - BAC FAIR SP.10 to M.Z.) and by the Italian Ministry of University and Research (PRIN grant n. C53D23004110006 to M.Z.). K.H. acknowledges the support of the China Scholarship Council (ID: 202307820031).

\section*{Author Contributions}
A.T. and M.Z. designed the research, acquired funding and supervised project development. A.T. drafted the manuscript. A.T. and K.H. carried out the experiments. K.H. produced the figures and analyzed the results. All authors revised and proof-read the final manuscript.

\section*{Competing Interests}
All authors declare no financial or non-financial competing interests.

\bibliographystyle{unsrt}  
\bibliography{NumSense} 
\newpage

\section*{Supplementary Information}

\setcounter{figure}{0}
\makeatletter 
\renewcommand{\thefigure}{S\@arabic\c@figure}
\makeatother

\subsection*{Estimation baselines}
\begin{figure}[h]
    \centering
    \includegraphics[width=0.55\linewidth]{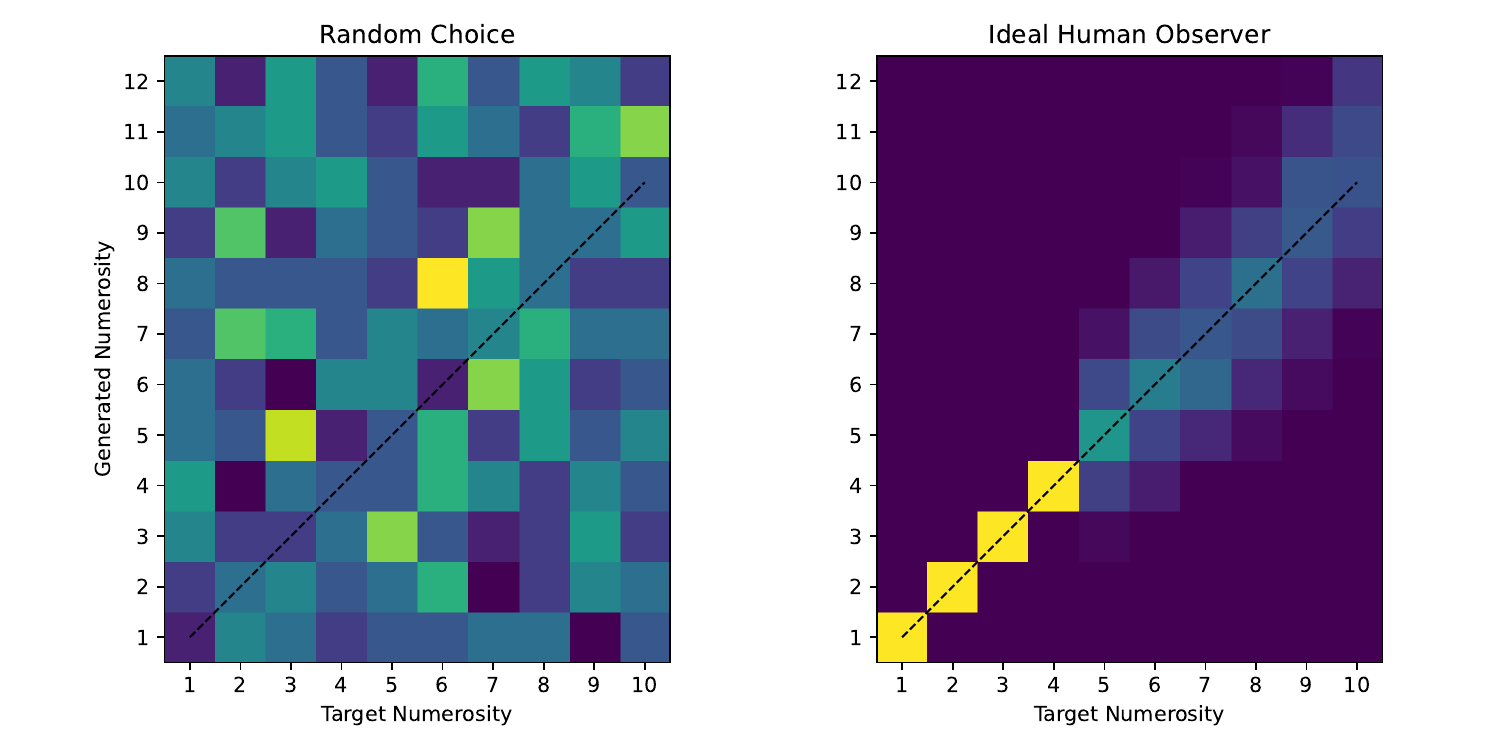}
    \caption{Confusion matrices for baseline models. On the left is the random choice model, while on the right is the ideal human observer, which has perfect accuracy in the subitizing range (1-4) and approximately estimates larger numbers according to Weber's law.}
    \label{fig:base_human_CM}
\end{figure}

The confusion matrices of the random model and the ideal human observer are shown in Figure \ref{fig:base_human_CM}.

\begin{figure}[h]
    \centering
    \includegraphics[width=0.95\linewidth]{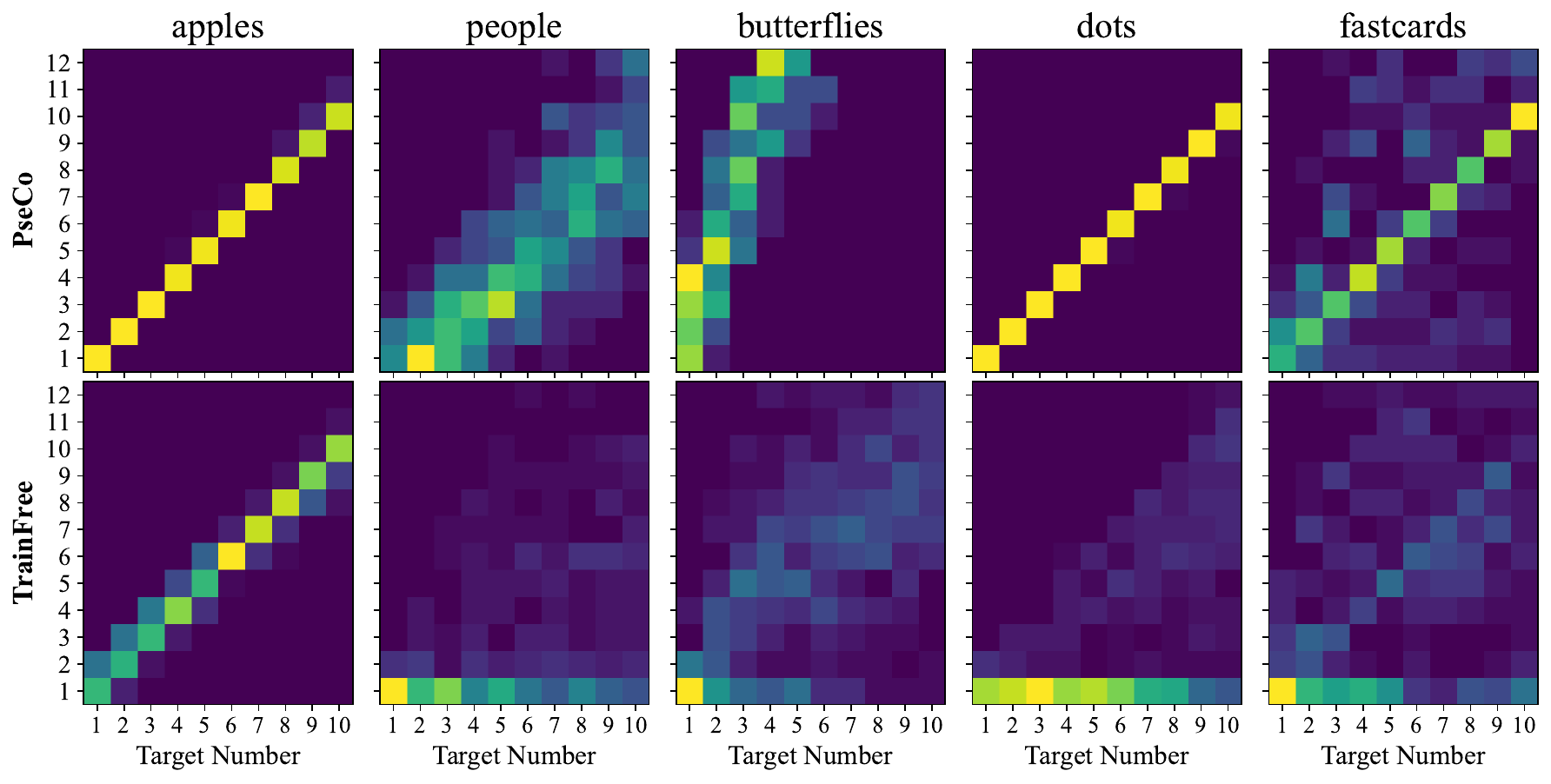}
    \caption{ {Confusion matrices for the counting-specific models for the numerosity naming task. Each panel shows the distribution of models' responses across different object categories: apples, people, butterflies, dots and fast cards. The x-axis represents the target number, while the y-axis represents the corresponding model responses. Response frequency is encoded using a perceptually uniform colormap (blue = 0\%, yellow = 100\%).}}
    \label{fig:7}
\end{figure}

 {
\subsection*{Counting-specific architectures}
The confusion matrices of the counting-specific architectures we considered as further benchmarks are shown in Figure \ref{fig:7}. Neither the PseCo nor the TrainFree models can reliably solve the numerosity naming task across all object categories.}

\begin{figure}[h]
    \centering
    \includegraphics[width=0.9\linewidth]{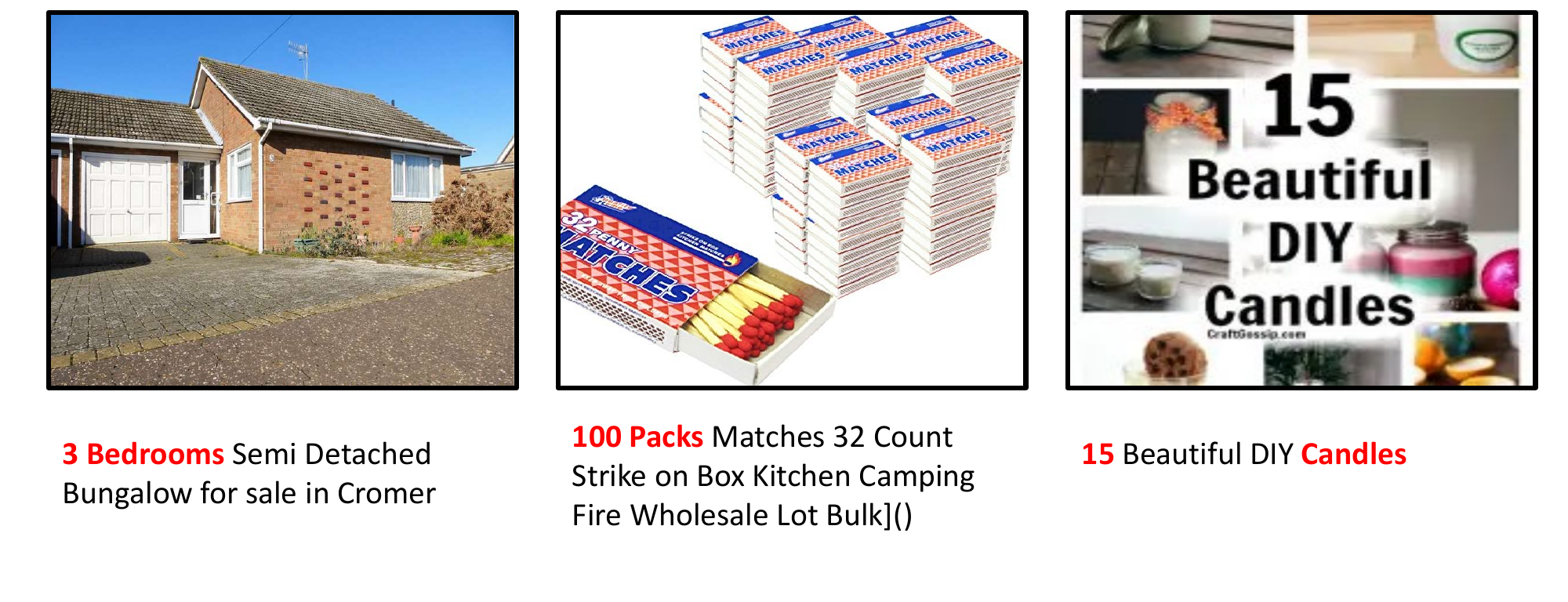}
    \caption{Examples of numerical misalignment between images and textual captions in the LAION-400M training corpora. The numerical information in the text is highlighted in red.}
    \label{fig:missalign}
\end{figure}

\subsection*{Automatic annotation of generated images}
\begin{figure}[h]
    \centering
    \includegraphics[width=0.8\linewidth]{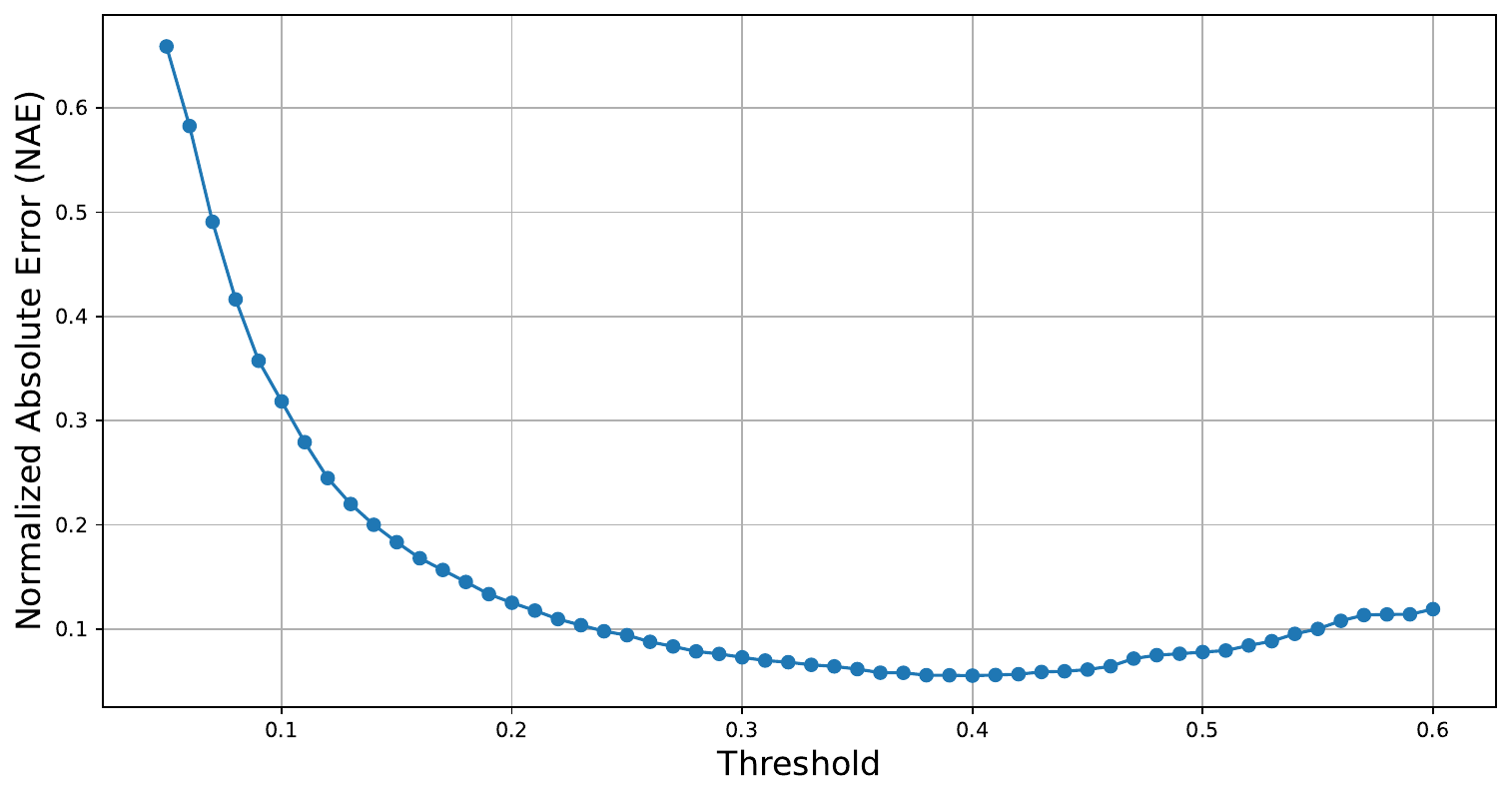}
    \caption{ {NAE of the image validation set as a function of Grounding DINO's threshold.}}
    \label{fig:8}
\end{figure}

The number of objects in each generated image was estimated using an advanced computer vision pipeline \cite{hou2025estimating}, validated against a separate set of 4,000 generated images that were manually labeled by one of the authors (K.H.) and independently verified by another author (A.T.). This pipeline employs Grounding DINO \cite{liu2025grounding}, a state-of-the-art object detection model, to identify objects within the generated images. The process begins with parsing the prompts used for the text-to-image models to identify the object category for each image. Grounding DINO then detects all objects belonging to the specified category by providing bounding boxes for each detected object, along with a confidence score for each bounding box. Bounding boxes with low confidence scores are filtered out.
To align the detections made by Grounding DINO with human annotations, a grid search was performed to optimize its confidence score threshold based on the NAE metric. The grid search explored confidence scores from 0.01 to 0.99 in increments of 0.01. The confidence score was iteratively adjusted to minimize the NAE between Grounding DINO's outputs and the human annotations. The optimal confidence score was determined to be 0.40, achieving a minimum NAE of 0.05.  {We provide the NAE as a function of different thresholds in Fig. \ref{fig:8}}. This calibration ensured that discrepancies between the object counts detected by Grounding DINO and the human annotations were minimized, thereby enhancing the accuracy and reliability of the automatic evaluation process.

\subsection*{Identifying countable numerosities from image-text pair datasets}\label{filter_criteria}
In order to identify target numerosities, we first defined a valid numerosity as a textual segment containing a numerical token followed by either a noun or an optional adjective plus a noun (e.g., ``1 apple'' or ``3 tall trees''). This criterion allowed to focus on explicit numerical references to countable objects while excluding textual references where numbers serve as model identifiers (e.g., ``Porsche 911'') or cardinal descriptors in product names (e.g., ``35th anniversary''). To implement this filtering process, we employed the \href{https://spacy.io/}{spaCy} library to determine the part-of-speech (POS) tags for each word in the captions. After locating candidate structures, we systematically excluded measurements and units. The exclusion criteria encompassed a comprehensive range of metric prefixes, spanning from the microscopic to the massive: micro-, milli-, centi-, kilo-, mega-, giga-, nano-, tera-, and peta-. These prefixes were filtered out when combined with common measurement units. The excluded base units covered fundamental physical quantities: distance (meter/metre, mile, foot, inch, yard), volume (liter/litre), mass (gram, ounce, pound), power (watt), electrical potential (volt), current (amp), and energy (joule). Time-related measurements were similarly excluded, ranging from seconds to years, as were temperature scales (Fahrenheit, Celsius). By applying filtering criteria, we derived a more precise set of image-text pairs that genuinely reflected countable numerosities.

\end{document}